# TextureMeDefect: LLM-based Defect Texture Generation for Railway Components on Mobile Devices

Rahatara Ferdousi
School of Electrical Engineering and Computer Science, University of Ottawa, Ontario, Canada. Email: rferd068@uottawa.ca
M. Anwar Hossain
School of Computing, Queen's University, Kingston, Ontario, Canada. Email: ahossain@queensu.ca
Abdulmotaleb El Saddik
School of Electrical Engineering and Computer Science, University of Ottawa, Ontario, Canada. Email: elsaddik@uottawa.ca

*Abstract*—Texture image generation has been studied for various applications, including gaming and entertainment. However, context-specific realistic texture generation for industrial applications, such as generating defect textures on railway components, remains unexplored. A mobile-friendly, LLM-based tool that generates fine-grained defect characteristics offers a solution to the challenge of understanding the impact of defects from actual occurrences. We introduce TextureMeDefect, an innovative tool leveraging an LLM-based AI-Inferencing engine. The tool allows users to create realistic defect textures interactively on images of railway components taken with smartphones or tablets. We conducted a multifaceted evaluation to assess the relevance of the generated texture, time, and cost in using this tool on iOS and Android platforms. We also analyzed the software usability score (SUS) across three scenarios. TextureMeDefect outperformed traditional image generation tools by generating meaningful textures faster, showcasing the potential of AI-driven mobile applications on consumer-grade devices.

*Index Terms*—AI, LLM, Mobile Device, AI-tool, Defect, Railway, Texture.

## I. INTRODUCTION

Texture image generation is not an innovation, especially in well-established domains such as gaming, art, and product design [1]. However, the generation of fine-grained, specific textures remains an underexplored field. Defect texture generation, in particular, is a domain-specific task that requires capturing subtle details rather than regular texture patterns [2]. In industrial inspection fields, such as railway infrastructure inspection, defect texture generation is essential for accurately understanding defects and preventing potential hazards [3]. These tasks demand the ability to capture fine-grained defect information (e.g., size, orientation, color, etc.) while providing a user-friendly, flexible solution for near real-time [4], on-site texture generation [5], without causing actual damage or manually designing defect textures, which can be time-consuming.

Existing methods for generating defect textures struggle to balance accuracy, flexibility, and practicality, especially on mobile devices [6]. Stochastic algorithms [7] [8] often fail because they rely on random processes that cannot capture complex, unusual texture patterns. Traditional AI-based tools also fall short, producing unrealistic defect textures due to hallucination and lack of context [3]. This challenge is particularly significant for rail components, where accurate visual descriptions of defect textures are crucial [9].

To address these gaps, we present TextureMeDefect [1], a cross-platform web tool that leverages LLMs to generate realistic defect textures directly on mobile devices. A base LLM is fine-tuned using a high-quality synthetic dataset generated using our proposed pipeline. Then the fine-tined model refines user-provided prompts, ensuring they accurately encapsulate defect characteristics like location, size, and orientation. These refined prompts are fed into a text-to-image generation model, and the resulting texture undergoes further processing for seamless integration with 3D models for visualization.

Through rigorous evaluation on a range of mobile devices, we assess the quality and relevance (accuracy) of the generated textures; the latency of the generation process, and the overall user experience. Our results demonstrate that TextureMeDefect surpasses existing text-to-image generation models in terms of consistency and accuracy, attributed to our innovative LLM-based approach. Furthermore, the software usability score (SUS) underscores the effectiveness and limitations of the proposed tool across three different scenarios.

The key contributions of this paper are three folds: i) to design and develop a synthetic defect texture generation pipeline [2], ii) The creation of a multimodal instruction-following defect dataset which captures the intricate details of various defects. iii) The design and development of TextureMeDefect [3], an LLM-based tool for generating realistic defect textures on mobile devices.

The remainder of this paper is organized as follows. Section II provides a summary of related work in texture generation and AI applications for industrial inspections. Section III elaborates on the methodology behind TextureMeDefect, in-cluding the dataset creation, model fine-tuning, and texture generation pipeline. Section IV presents the experimental setup and results of our multifaceted evaluation. Finally, Section V concludes the paper with a discussion of the implications of our work and future research directions.

---
[1]Click to watch the promo
[2]Click to use the code and create your own customized dataset
[3]Click to download the dataset.

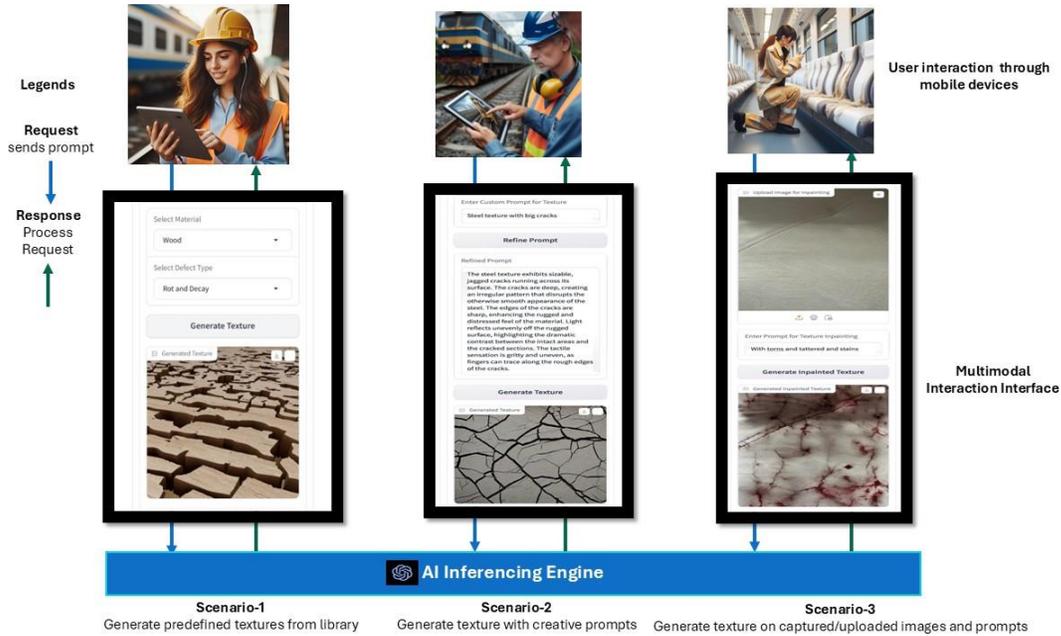

Fig. 1: System Architecture of TextureMeDefect: An LLM-Based Tool for Defect Texture Generation on Mobile Devices

## II. RELATED WORK

In this section, we outline the work focused on texture generation. As there are no specific tools to generate defect texture, we also studied related work to generate images using LLM models.

### A. Texture Generation in Literature

Traditional approaches often fall short of comprehensively addressing these combined requirements [10]. Non-AI-based methods, such as procedural modeling and image manipulation techniques, can be computationally expensive and lack the adaptability to handle a wide variety of defect types and component geometries [8].

### B. AI-based tools for image generation

Although AI-based techniques show promise, they often rely on limited or domain-specific datasets, hindering their generalizability and accuracy [11]. Current multimodal LLMs, such as DALLE-3 [12], InstructPix2Pix [13], and SDXL [14], have shown promise in various domains, including healthcare, and creative content generation [15]. However, tools using these models often struggle to capture the fine-grained, context-specific details required for domain-specific image generation [16] [17]. For instance, a common defect image like a 'crack on the rail track' might be generated accurately by GPT-4, but creating the crack texture in detail is not as precise. For training or simulation purposes, such textures are extremely important in modeling defect characteristics on a 3D model [18]. Additionally, traditional image generation tools are computationally intensive [6], rendering them impractical for use on resource-constrained mobile devices [15] in terms of texture generation time and cost-per-token.

### C. Gaps and Requirements

Based on the above study, we conclude that while there are efforts in the literature for surface or body texture generation, the evaluation of a texture generation tool on mobile devices, as well as custom domain texture generation like defect texture generation on railway components, is a rare use case to explore.

## III. METHODS

The system architecture depicted in Fig. 2, showcases the design of TextureMeDefect. TextureMeDefect is designed as a multimodal defect texture generation tool to be deployed in mobile devices like-smartphone or tablet. The breakdown of the architecture along with the details of the proposed AI-inferencing engine, is in the following sections.

### A. Multimodal User Interaction Interface (MUII)

The system is designed to be interactive, allowing users to generate defect textures through a multimodal interface. The user can interact with the system via mobile devices through gesture, touch, and keyboard input. The user sends a request to the AI-inferencing engine and consumes the output that the AI-inferencing engine produces by processing the request through this MUII. Let us understand the role of the MUII in TextureMeDefect through the following example scenarios.

In Scenario-1 (Library-Based Selection), users select predefined materials and defects from a library to generate textures using pre-trained models; in Scenario-2 (Creative Prompt-based Generation), users write prompts to describe desired textures, which the system refines and generates; and in Scenario-3 (Image-Based Generation), users upload images and specify defects to visualize them on the material, enabling a simulative understanding of defects on real-life components.

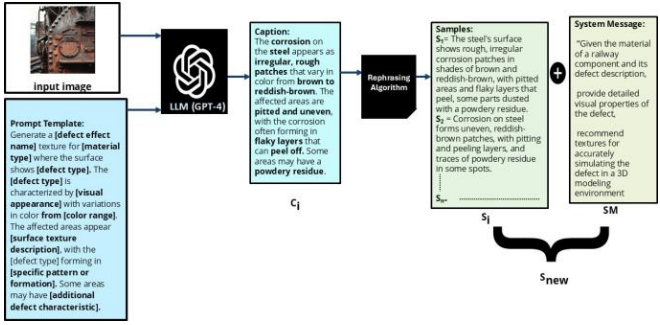

Fig. 2: Multimodal Instruction Following Synthetic Dataset Generation Pipeline for Defect Texture

### B. AI Inferencing Engine

At the core of the system is the AI inferencing engine, which processes all user inputs and requests for defect generation. It operates in the following steps:

1) **Step-1(Synthetic Dataset Generation):** We leverage the capabilities of GPT-4 to generate a synthetic dataset consisting of image-caption pairs. The proposed dataset generation pipeline processes an input image and uses a prompt template designed to make GPT-4 capture the visual characteristics of defect textures. These visual components are selected based on existing literature on railway defect inspection and texture generation [9]. The generated captions are then passed to Algorithm 1, which has already been applied in one of our ongoing research projects [18]. The algorithm produces concise and diverse samples for each caption and aggregates them with a corresponding system message. By compiling these samples into a JSON file, we create the defect texture multimodal instruction-following dataset. Since defect samples in railway components are rare, the algorithm generates a variety of samples for each caption, ensuring sufficient data for fine-tuning a customized LLM specialized in defect texture-related knowledge.

2) **Step 2 (Base LLM Fine-tuning):** We fine-tune a GPT-3 model on the defect texture focused synthetic dataset, following the principle that models trained on data generated or revised by similar models tend to achieve superior performance [11]. Fine-tuning is performed iteratively, with multiple passes over the dataset to ensure that the model adapts to the variety of defect detection.

3) **Step 3 (Prompt Tuning):** User-provided prompts, whether textual descriptions or annotations on captured images, are refined using the fine-tuned GPT-3 model. This ensures that the prompts accurately capture the de-

**Algorithm 1** Texture Description Rephrasing Algorithm

**Input:** A list of texture-based captions from defect images: `captions = [`$c_1$`, `$c_2$`, ..., `$c_n$`]`

**Output:** A dataset with diverse texture descriptions per caption, each associated with a system message.

**Step-1:** Create an empty list: `TextureDS = []`

**Step-2:** For each texture caption ($c_i$) in `captions`, create an empty set for unique rephrased descriptions: `TextureSamples = {}`

**Step-3:**

**while** the number of texture samples is less than $K$ **do**
    Generate a new texture description ($t_{new}$) using a language model based on $c_i$.
    Add $t_{new}$ to `TextureSamples`.
**end while**

**Step-4:** For each generated texture description ($t_{new}$) in `TextureSamples`, formulate a system message describing its generation process or characteristics.
Combine $t_{new}$ with the system message to form a structured data entry.

**Step-5:** Append all structured entries from `TextureSamples` to `TextureDS`. Ensure `TextureDS` does not contain duplicates.

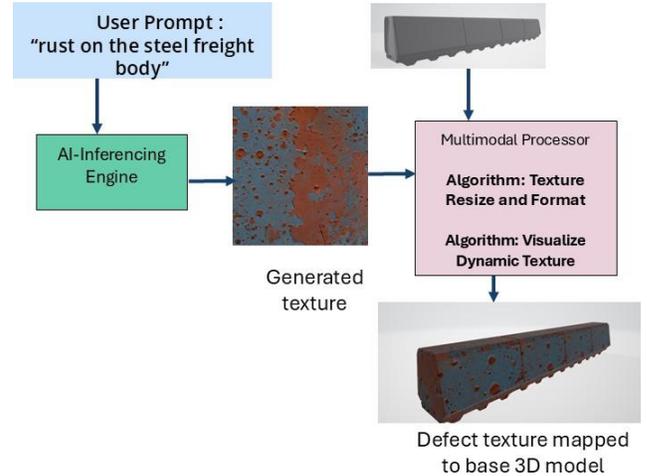

Fig. 3: Multimodal Instruction Following Synthetic Dataset Generation Pipeline for Defect Texture

sired defect characteristics, leading to more precise and relevant texture generation. For Example: User Input: "crack on the rail" Tuned Prompt: "A transverse crack, approximately 2 inches long, located on the head of the rail, with slight rust discoloration around the edges." By generating the specific prompt we target to capture the

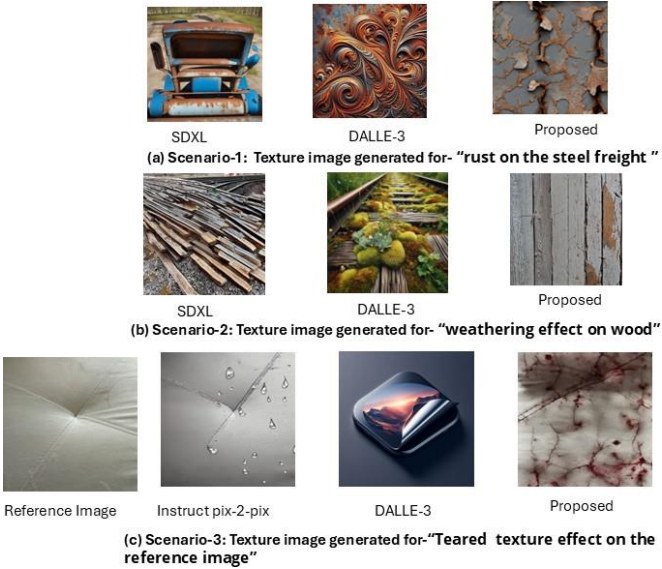

Fig. 4: Multimodal Processor for Formatting the Generated Texture into User's Consumable Format

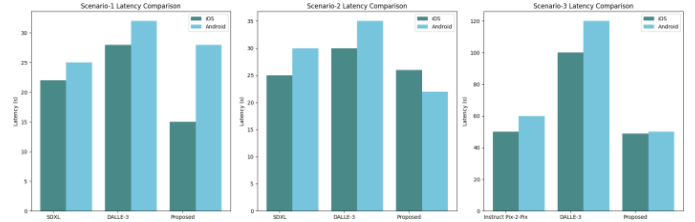

Fig. 5: Latency of Texture Image Generation for three scenarios. Scenario-1: Predefined prompt, Scenario-2: Custom Prompt and Scenario-3: Inpaint Prompt

visual details of the texture while prompting an image generation model.

4) **Step 4 (Multimodal Image Generation):** The multimodal instruction following dataset enables the image generation model to create defect textures by accurately capturing essential texture components such as cracks, wear, or decay. In this research, we adopt the Multimodel agent concept, where a fine-tuned LLM is used to interface with multiple image generation models. For example, to handle Scenario-1 and Scenario-2 in Fig.2, we use the SDXL model for text-to-image generation. Meanwhile, for Scenario-3 in Fig.2, we utilize the InstructPix2Pix model, which can process both image and text inputs.

5) **Step 5 (Multimodal Processing):** Once a defect texture is generated, it may not always be ready for direct application to a 3D model. The system includes image-processing algorithms for resizing and scaling, converting the textures into standardized formats for 3D model integration as depicted in Fig. 3. The final processed texture is displayed on the user's mobile device, allowing them to visualize the simulated defect on the captured image or apply it to a 3D model for further inspection or analysis.

## IV. EXPERIMENT AND RESULTS

We conducted a multifaceted evaluation of the proposed approach for the TextureMeDefect tool. The following are the details of the experimental procedures and findings.

### A. Experiment-1: Relevance of Output

We compared the qualitative output of the proposed model with similar models. It is to be noted that the other models were also accessed via an interface through Replicate Playground. So that the impact of our designed interface for human-ai interaction can be understood.

As depicted in Fig. 4, for all three scenarios, our proposed tool TextureMeDefect captured the fine-grained details to address the user's request more appropriately than other models. For scenario-1, the base model SDXL generated texture on a small freight, while DALLE-3 generated creative art on the texture. Because none of these models are customized for defect texture details specifically, this makes these textures unrealistic to apply directly to a 3D model. A similar result is seen for the custom prompt-based texture generation. Interestingly, for scenario-3, the instruct pix-2-pix base model generated an image with a tear (in the context of crying) due to hallucination. Additionally, for such advanced features to generate in-paint texture, DALLE-3 completely fails to produce relevant output. This is where our proposed AI-inferencing Engine, stands out compared to the existing solutions by generating realistic textures.

### B. Experiment-2: Latency

We evaluated the latency of the texture image generation across the three scenarios and compared the performance with existing solution on iOS and Android platform. We employed a 10th Generation Ipad and Samsung Galaxy S24 phone during the evaluation. The results from this section onwards have been computed as the mean of 50 image generations per scenario, totaling 150 texture images.

As depicted in Fig. 5, In Scenario 1, the proposed model outperforms SDXL and DALLE-3 on both iOS and Android. It has a latency of 15-20 seconds on iOS and 18-30 seconds on Android, compared to SDXL (22 seconds) and DALLE-3 (28 seconds) on iOS, and 25 and 32 seconds on Android, respectively. In Scenario 2, the proposed model also leads with latencies of 18-30 seconds on both platforms. SDXL and DALLE-3 show higher latencies, especially on Android (30-35 seconds) compared to iOS (25-30 seconds). In Scenario 3, involving intricate image modifications, the proposed model performs best with latencies of 40-50 seconds on iOS and 50-60 seconds on Android. Instruct Pix-2-Pix has higher latencies (50-100 seconds on iOS and 60-120 seconds on Android), while DALLE-3 lags significantly (100 seconds on iOS and up to 120 seconds on Android). The complexity of tasks in

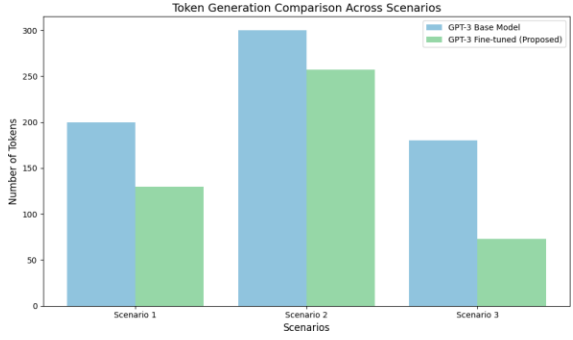

Fig. 6: Number of Generated Tokens for Tuning the prompt Base vs Fine-Tuned GPT-3 Model

Scenario 3 increases computational demands, making the proposed model the most efficient. iOS consistently outperforms Android in latency across all three scenarios, with Android experiencing up to a 30% increase in latency. Although the Samsung Galaxy S24 features powerful hardware, Android's more fragmented ecosystem and reliance on GPU processing may introduce additional overhead, leading to slower performance.

*C. Experiment-3: Token*

It can be observed form Fig. 6 The base model generated around 180 tokens for Scenario-1, while the fine-tuned model generated significantly fewer tokens, around 50-150 tokens.This large reduction can be attributed to the prompt tuning of the fine-tuned model, which is better equipped to generate concise, relevant responses. Since the proposed model was fine-tuned specifically for defect texture understanding, it filters out unnecessary information, leading to more efficient token usage. In Scenario 2, the complexity and variability of user prompts led to a higher token count for both models. However, the fine-tuned model still outperformed the base model, especially in handling prompts with extra information. Like Scenario 1, the fine-tuned model's performance in Scenario 3 was markedly better, reducing the number of tokens substantially.

*D. Estimated Cost Analysis*

We estimate a cost analysis based on both token generation and processing time for each scenario using the fine-tuned model. As illustrated in the graphs in Fig. 5, 6, cost depends on these two factors, which vary significantly across the scenarios. We estimate the cost using the following formula:

$$C = \sum_{i=1}^{n} (\text{Tokens}_i \times \text{Cost per Token}_i) + \sum_{j=1}^{m} \text{Processing Time}_j \times \text{Cost per Second}_j \quad (1)$$

Where $\text{Tokens}_i$ is the number of tokens generated in the $i^{th}$ task. Cost per $\text{Token}_i$ is the cost per token in the $i^{th}$ task. Processing $\text{Time}_j$ is the processing time in seconds for the $j^{th}$ model. Cost per $\text{Second}_j$ is the processing cost per second in the $j^{th}$ model.

In Scenario-1, token generation is low (50-150 tokens) with a short processing time, resulting in the lowest overall cost. This makes it highly cost-efficient. In Scenario-2, token generation increases to 100-260 tokens due to more complex prompts, raising the overall cost moderately despite manageable processing times. Scenario-3, although having a low token count (50-150 tokens), has significantly longer processing times (40-260 seconds). This extended duration increases the cost substantially, making it the most expensive scenario despite the lower token count.

The analysis indicates that while token generation is a key factor, processing time plays a critical role in overall cost, particularly for scenarios involving more computationally demanding tasks like Scenario-3. Optimizing both token efficiency and processing speed is essential for reducing costs in large-scale applications.

*E. Usefuleness*

We conducted a software usability score (SUS) testing for each of the three scenarios aimed at defect texture image generation as follows.

*1) Participants:* A total of 15 voluntary participants took part in the testing with following distribution-Android users: 46.67%, iOS (iPad) users: 53.33%. 87.10% of the users are expert users who are familiar with generating image using AI-tools. Non-expert users are only 12.90% who have never peformed AI-based image generation.

*2) Procedure:* The objective is to evaluate how users interact with TextureMeDefect through three scenarios stated earlier. Here are the modified 10 SUS questions to evalaute the TextureMeDefect tool, focusing on the defect texture generation process: i) **Q1:**I found the process of generating defect textures using the input (dropdown/textbox/image upload button) options to be straightforward. ii)**Q2:**I felt confident generating defect textures by using this user interface. iii)**Q3:**The system's instructions for generating textures through different modes (dropdown, prompt, image upload) were clear and easy to understand. iv)**Q4:** I believe that most users would learn to use this tool quickly. v)**Q5:**I felt that the time taken to generate defect textures was reasonable. vi)**Q6:** I found the system unnecessarily complex when using custom prompts. (Reverse scored) vii)**Q7:** I would need technical assistance to use the tool for generating defect textures. (Reverse scored) viii)**Q8:**The tool was smooth and responsive when generating textures from images and descriptions. ix)**Q9:**I found the generated defect textures to be realistic and aligned with my input. x)**Q10:**I would use this tool again for defect texture generation in the future.

*3) Score interpretation:* According to Fig. 7, The System Usability Scale (SUS) evaluates usability on a scale from 0 to 100. In Scenario-1, the tool demonstrated good usability with an average score close to 70%, particularly excelling in ease of use (Q1) and time taken (Q5), both rated 4.9, along with high confidence (Q2: 4.8) and realism of textures (Q9: 4.9). Scenario 2 scored lower, with an average of 55%, indicating issues with instruction clarity (Q3: 3.5) and tool complexity. Scenario-3, like Scenario-1, had a 70% average, with users rating ease of use (Q1: 4.9), time taken (Q5: 4.7), and confidence (Q2: 4.7) positively. However, complexity (Q6, reverse scored) and need for assistance (Q7, reverse scored) were slightly lower, suggesting some difficulty with advanced inputs.

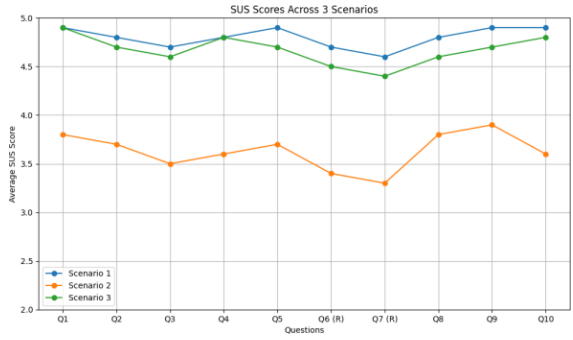

Fig. 7: Average SUS score per question across three scenarios

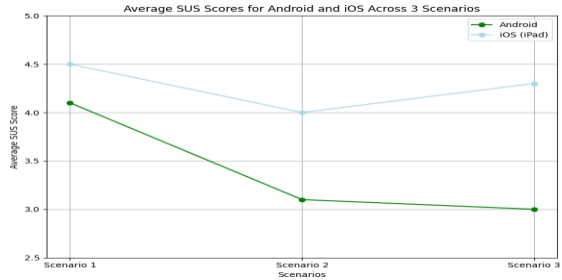

Fig. 8: Average SUS score per question across three scenarios on different platforms

Similar to the token generation in Fig. 6, iOS users consistently rated the system higher across all scenarios, with a slight dip in Scenario-2 but a recovery in Scenario-3 due to better output quality despite the extended time (See Fig. 8. Android users experienced a noticeable decline in usability from Scenario 1 to Scenario 3, reflecting more challenges with input options and longer times to generate realistic outputs.

## V. CONCLUSION AND FUTURE WORK

In this work, we designed, developed, and evaluated the first defect texture generation tool for mobile devices, TextureMeDefect. We outperformed traditional image generation tools by following the best practices in designing the AI-inferencing engine. Our experiments demonstrate that TextureMeDefect generates relevant and useful outputs in a shorter time compared to existing tools, though there is still room for improvement. The usability scores and quantitative results indicate that the custom prompt-based texture generation interface needs to be more user-friendly, as some participants found it difficult to generate textures after entering their own prompts. This step could be simplified in the interface and the latency for the inpaint texture image generation could be reduced in the future extension of this work.